\documentclass[sigconf,nonacm]{acmart}
\usepackage{color}
\newtheorem{definition}{Definition}
\newtheorem{theorem}{Theorem}
\newtheorem{proposition}{Proposition}
\newtheorem{axiom}{Axiom}

\usepackage[ruled,vlined]{algorithm2e}
\usepackage{bm}
\usepackage{wrapfig}
\usepackage{graphicx}
\usepackage{subfigure}
\usepackage{enumitem}
\usepackage{multirow}
\usepackage{amsthm,amsmath}
\usepackage{setspace}
\definecolor{text}{rgb}{0.8,0.1,0.1}
\newcommand{\sen}[1]{{\color{black}{#1}}}
\newcommand{\cui}[1]{{\color{black}{#1}}}
\newcommand{\liang}[1]{{\color{black}{#1}}}

\AtBeginDocument{%
  \providecommand\BibTeX{{%
    \normalfont B\kern-0.5em{\scshape i\kern-0.25em b}\kern-0.8em\TeX}}}

\copyrightyear{2022}
\acmYear{2022}
\setcopyright{acmcopyright}
\acmConference[KDD '22] {Proceedings of the 28th ACM SIGKDD Conference on Knowledge Discovery and Data Mining}{August 14--18, 2022}{Washington, DC, USA.}
\acmBooktitle{Proceedings of the 28th ACM SIGKDD Conference on Knowledge Discovery and Data Mining (KDD '22), August 14--18, 2022, Washington, DC, USA}
\acmPrice{15.00}
\acmISBN{978-1-4503-9385-0/22/08}
\acmDOI{10.1145/3534678.3539237}

\begin{document}

\title{Collaboration Equilibrium in Federated Learning}

\author{Sen Cui$^{1}$, Jian Liang$^{2}$, Weishen Pan$^{1}$, Kun Chen$^{3}$, Changshui Zhang$^{1}$, Fei Wang$^4$}
\affiliation{%
  \institution{$^1$Institute for Artificial Intelligence, Tsinghua University (THUAI), State Key Lab of Intelligent Technologies and Systems,Beijing National Research Center for Information Science and Technology (BNRist) \\ Department of Automation, Tsinghua University, Beijing, P.R.China\\$^2$Alibaba Group, China\\$^3$Department of Statistics, University of Connecticut, USA\\$^4$Department of Population Health Sciences, Weill Cornell Medicine, USA}
  \country{}
}

\email{cuis19@mails.tsinghua.edu.cn, xuelang.lj@alibaba-inc.com,pws15@mails.tsinghua.edu.cn,kun.chen@uconn.edu, zcs@mail.tsinghua.edu.cn,few2001@med.cornell.edu}

\begin{abstract}
Federated learning (FL) refers to the paradigm of learning models over a collaborative research network involving multiple clients without sacrificing privacy. Recently, there have been rising concerns on the distributional discrepancies across different clients, which could even cause counterproductive consequences when collaborating with others. While it is not necessarily that collaborating with all clients will achieve the best performance, in this paper, we study a rational collaboration called ``collaboration equilibrium'' (CE), where smaller collaboration coalitions are formed. Each client collaborates with certain members who maximally improve the model learning and isolates the others who make little contribution. We propose the concept of benefit graph which describes how each client can benefit from collaborating with other clients and advance a Pareto optimization approach to identify the optimal collaborators. Then we theoretically prove that we can reach a CE from the benefit graph through an iterative graph operation. Our framework provides a new way of setting up collaborations in a research network. Experiments on both synthetic and real world data sets are provided to demonstrate the effectiveness of our method.
\end{abstract}


\maketitle
\section{Introduction}

Effective learning of machine learning models over a collaborative network of data clients has drawn considerable interest in recent years. Frequently, due to privacy concerns, we cannot simultaneously access the raw data residing on different clients. Therefore, distributed \citep{li2014scaling} or federated learning ~\citep{mcmahan2017communication} strategies have been proposed, where typically model parameters are updated locally at each client with its own data, and the parameter updates, such as gradients, are transmitted out and communicate with other clients. During this process, it is usually assumed that the participation in the network comes at no cost, i.e., every client is willing to participate in the collaboration. However, this is not always true in reality.

One example is the clinical research network (CRN) involving multiple hospitals \citep{fleurence2014launching}. Each hospital has its own patient population. The patient data are sensitive and cannot be shared with other hospitals. If we want to build a risk prediction model with the patient data within this network in a privacy-preserving way, the expectation from each hospital is that a better model can be obtained through participating in the CRN compared to the one built from its own data collected from various clinical practice with big efforts. In this scenario, there has been a prior study showing that the model performance can decrease when collaborating with hospitals with very distinct patient populations due to negative transfer induced by sample distribution discrepancies~\citep{wang2019characterizing,pan2009survey}.

With these considerations, in this paper, we propose a novel {\em learning to collaborate} framework. We allow the participating clients in a large collaborative network to form non-overlapping collaboration coalitions. Each coalition includes a subset of clients such that the collaboration among them can benefit their respective model performance. We aim at identifying the collaboration coalitions that can lead to a {\em collaboration equilibrium}, i.e., there are no other coalition settings that any of the individual clients can benefit more (i.e., achieve better model performance).

\begin{figure*}[h!]
    \centering{
    \includegraphics[width=2.0\columnwidth]{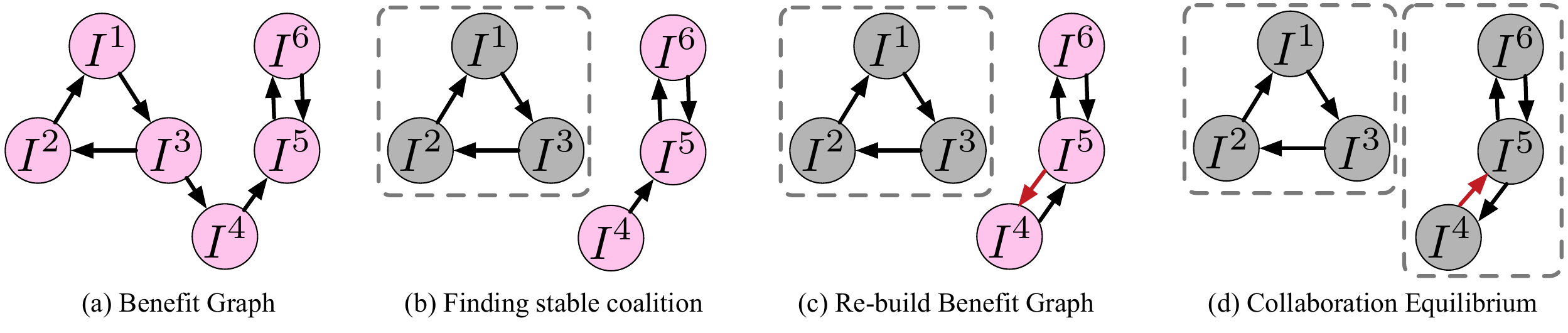}}
    \caption{(a) The benefit graph on all clients; (b) Finding all stable coalitions and remove them; (c) \sen{reconstruct the benefit graph on the remaining clients; after $I^{3}$ is removed, $I^{4}$ re-identifies its necessary collaborators in $\left\{ I^{4}, I^{5}, I^{6} \right\}$  which is $\{I^{5} \}$ as the added the red arrow from $I^{5}$ to $I^{4}$ in the figure}; (d) iterate (b) and (c) until achieving collaboration equilibrium.}
    \label{fig:intro}
\end{figure*}

To obtain the coalitions that can lead to a collaboration equilibrium, we propose a Pareto optimization framework to identify the necessary collaborators for each client in the network to achieve its maximum utility. In particular, we optimize a local model associated with a specific client on the Pareto front of the learning objectives of all clients. Through the analysis of the geometric location of such an optimal model on the Pareto front, we can identify the necessary collaborators of each client. The relationships between each client and its necessary collaborators can be encoded in a {\em benefit graph}, in which each node denotes a client and the edge from $I^{j}$ to $I^{i}$ represents $I^{j}$ is one of the necessary collaborators for $I^{i}$, as exemplified in Figure~\ref{fig:intro} (a). Then we can derive the coalitions corresponding to the collaboration equilibrium through an iterative process introduced as follows. Specifically, we define a stable coalition as the minimum set such that its all involved clients can achieve its maximal utility. From the perspective of graph theory, these stable coalitions are the strongly connected components of the benefit graph. For example, $C = \left\{I^{1}, I^{2}, I^{3} \right\}$ in Figure~\ref{fig:intro} (b) is a stable coalition as all clients can achieve their best performance by collaborating with the clients in $C$ (compared with collaborating with other clients in the network). By removing the stable coalitions and re-building the benefit graph of the remaining client iteratively as shown in Figure~\ref{fig:intro} (b) and (c), we can identify all coalitions as in Figure~\ref{fig:intro} (d) and prove that the obtained coalitions can lead to a collaboration equilibrium.

We empirically evaluate our method on synthetic data, UCI adult~\citep{kohavi1996scaling}, a classical FL benchmark data set CIFAR10~\citep{krizhevsky2009learning}, and a real-world electronic health record (EHR) data repository eICU~\citep{pollard2018eicu}, which includes patient EHR data in ICU from multiple hospitals. The results show our method significantly outperforms existing relevant methods. The experiments on eICU data demonstrate that our algorithm is able to derive a good collaboration strategy for the hospitals to collaborate. The source codes are made publicly available at \url{https://github.com/cuis15/learning-to-collaborate}.

\section{Related Work}
\subsection{Federated Learning}
Federated learning has raised several concerns, including communication efficiency~\cite{konevcny2016federated}, fairness~\cite{cui2021addressing,cui2021towards,pan2021explaining}, and statistical heterogeneity~\cite{karimireddy2020scaffold}, and they have been the topic of multiple research efforts~\cite{mohri2019agnostic}. In a typical FL setting, a global model~\cite{deng2020distributionally,mohri2019agnostic,reisizadeh2020robust,diamandis2021wasserstein,li2019fair} is learned from the data residing in multiple distinct local clients. However, a single global model may lead to performance degradation on certain clients due to data heterogeneity. Personalized federated learning (PFL)~\citep{kulkarni2020survey}, which aims at learning a customized model for each client in the federation, has been proposed to tackle this challenge. For example, ~\citet{zhang2020personalized} proposes to adjust the weight of the objectives corresponding to all clients dynamically; ~\citet{fallah2020personalized} proposes a meta-learning based method for achieving an effectively shared initialization of all local models followed by a fine-tuning procedure; ~\citet{shamsian2021personalized} proposes to learn a central hypernetwork which can generate a set of customized models for each client. FL assumes all clients are willing to participate in the collaboration and existing methods have not considered whether the collaboration can really benefit each client or not. Without benefit, a local client could be reluctant to participate in the collaboration, which is a realistic scenario we investigate in this paper. One specific FL setup that is relevant to our work is clustered federated learning~\citep{sattler2020clustered, mansour2020three}, which groups the clients with similar data distributions and trains a model for each client group. The scenario we are considering in this paper is to form collaboration coalitions based on the performance gain each client can get for its corresponding model, rather than sample distribution similarities.

\subsection{Multi-Task Learning and Negative Transfer}
Multi-task learning~\citep{caruana1997multitask} (MTL) aims at learning shared knowledge across multiple inter-related tasks for mutual benefits.  Typical examples include hard model parameter sharing~\citep{kokkinos2017ubernet}, soft parameter sharing~\citep{lu2017fully}, and neural architecture search (NAS) for a shared model architecture~\citep{real2019regularized}. However, sharing representations of model structures cannot guarantee model performance gain due to the existence of negative transfer, while we directly consider forming collaboration coalitions according to individual model performance benefits. In addition, MTL usually assumes the data from all tasks are accessible, while our goal is to learn a personalized model for each client through collaborating with other clients without directly accessing their raw data. It is worth mentioning that there are also clustered MTL approaches~\citep{standley2020tasks,zamir2018taskonomy} which assume the models for the tasks within the same group are similar to each other, while we want the clients within each coalition can benefit each other through collaboration when learning their respective models.


\subsection{Cooperative Game Theory}
Cooperative game theory studies the game with competition between groups of players, which focuses on predicting which coalitions will form. There are theoretical research~\cite{donahue2021model,donahue2021optimality} focusing on the linear regression and mean estimation problems in federated learning from a coalitional game perspective. Classical cooperative game theory~\cite{arkin2009geometric,aumann1974cooperative,yi1997stable} requires a predefined payoff function, so that it obtains the payoff of each group of players directly. Given a predefined payoff function over all possible groups, the payoff of each coalition is transferable among players in the coalition.

In our work, we study how to collaborate among clients to learn personalized models in federated learning. Suppose we analogize the model performance of each client to the payoff in cooperative game theory, this ``payoff function'' is not predefined and has to be approximated by the evaluation of the learned models on each client. Meanwhile, the performance of learned models is not transferable, so prior methods in cooperative game theory may not be used to reach a collaboration equilibrium directly.

\section{Collaboration Learning Problem}

We first introduce the necessary notations and definitions in Section \ref{sec:3-1}, and then define the collaboration equilibrium we aim to achieve in Section \ref{sec:3-2}.


\subsection{Definitions and Notations}
\label{sec:3-1}
Suppose there are $N$ clients $\bm{I} = \left\{I^{i}\right\}_{i=1}^{N}$ in a collaborative network and each client is associated with a specific learning task $T^{i}$ based on its own data $D^{i} = \left\{ X^{i}, Y^{i}\right\}, i \in \left\{ 1, 2,..., N\right\}$, where the input space $X^{i}$ and the output space $Y^{i}$ may or may not share across all $N$ clients. Each client pursues collaboration with others to learn a personalized model $M^{i}$ by maximizing its utility (i.e., model performance) without sacrificing data privacy. There is no guarantee that one client can always benefit from the collaboration with others, and the client would be reluctant to participate in the collaboration if there is no benefit. In the following, we describe this through a concrete example.

\paragraph{No benefit, no collaboration.}
\label{example1}
Suppose the local data $\{D^{i}\}, i \in \left\{ 1, 2,..., N\right\}$ owned by different clients satisfy the following conditions: 1) all local data are from the same distribution $D^{i} \sim \mathcal{P}$; 2) $D^{1} \subset D^{2} \subset D^{3} ,..., \subset D^{N}$. Since $D^{N}$ contains more data than other clients, $I^{N}$ cannot benefit from collaboration with any other clients, so $I^{N}$ will learn a local model using its own data. Once $I^{N}$ refuses to collaboration, $I^{N-1}$ will also work on its own as $I^{N-1}$ can only improve its utility by collaborating with $I^{N}$. $I^{N-2}$ will learn individually out of the same concerns. Finally, there is no collaboration among any clients.

Due to the discrepancies of the sample distributions across different clients, the best local model for a specific client is very likely to come from collaborating with a subset of clients rather than all of them. Suppose $U(I^{i}, C)$ denotes the model utility of client $I^{i}$ when collaborating with the clients in client set $C$. In the following, we define $U_{max}(I^{i}, C)$ as the maximum model utility that $I^{i}$ can achieve when collaborating with different subsets of $C$.
\begin{definition}[Maximum Achievable Utility (MAU)]
\label{def:mau}
This is the maximum model utility for a specific client $I^{i}$ to collaborate with different subsets of client set $C$:
$U_{max}(I^{i}, C) =  \max_{C' \subset C} U(I^{i}, C')$.
\end{definition}
\vspace{-.1cm}
From Definition \ref{def:mau}, MAU satisfies $U_{max}(I^{i}, C') \leq U_{max}(I^{i}, C)$ if $C'$ is a subset of $C$. Each client $I^{i} \in \bm{I}$ aims to identify its ``optimal set" of collaborators from $\bm{I}$ to maximize its local utility, which is defined as follows. 

\begin{definition}[Optimal Collaborator Set (OCS)]
\label{def:occ}
A client set $C^{opt}_{\bm{I}}(I^{i})\subset\bm{I}$ is an optimal collaborator set for $I^{i}$ if and only if $C^{opt}_{\bm{I}}(I^{i})$ satisfies
\begin{subequations}
\begin{align}
& \forall C \subset \bm{I}, \ U(I^{i}, C^{opt}_{\bm{I}}(I^{i})) \geq U(I^{i}, C); \label{eq:occ1} \\
& \forall C' \varsubsetneqq C^{opt}_{\bm{I}}(I^{i}), \ U(I^{i}, C^{opt}_{\bm{I}}(I^{i})) > U(I^{i}, C'). \label{eq:occ2}
\end{align}
\end{subequations}
\end{definition}
Eq.(\ref{eq:occ1}) means that $I^{i}$ can achieve its maximal utility when collaborating with $C^{opt}_{\bm{I}}(I^{i})$ and Eq.(\ref{eq:occ2}) means that all clients in $C^{opt}_{\bm{I}}(I^{i})$ are necessary. In this way, the relationships between any client $I^{i}\in\bm{I}$ and its optimal collaborator set $C^{opt}_{\bm{I}}(I^{i})$ can be represented by a graph which is called the \emph{benefit graph} (BG). Specifically, for a given client set $C\subset\bm{I}$, we use $BG(C)$ to denote its corresponding BG. For the example in Figure~\ref{fig:intro} (a), an arrow from $I^{j}$ to $I^{i}$ means $I^{j} \in C^{opt}_{\bm{I}}(I^{i})$, e.g., $I^{1}\rightarrow I^{3}$ means $I^{1} \in C^{opt}_{\bm{I}}(I^{3})$. For a client set $C$, if every member can achieve its maximum model utility through the collaboration with other members within $C$ (without collaboration with other members outside $C$), then we call $C$ a {\em coalition}.




\begin{figure}[h!]
    \centering{
    \includegraphics[width=0.7\columnwidth]{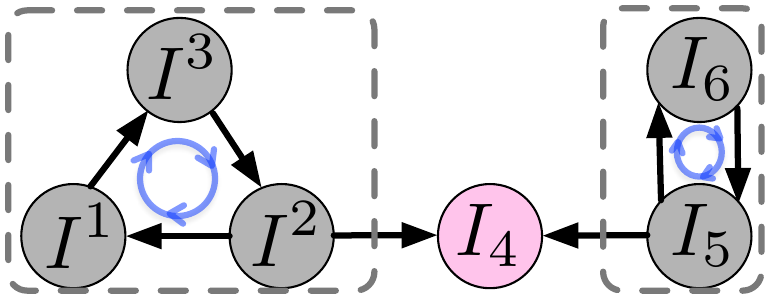}}
    \caption{Forming coalitions for maximizing the local utility.}
    \label{fig:example2}
\end{figure}

\paragraph{Forming coalitions for maximizing the local model utilities}
Figure~\ref{fig:example2} shows an example BG with 6 clients. 
$I^{3}$ can achieve its optimal model utility by collaborating with $I^{1}$. Similarly, $I^{1}$ and $I^{2}$ can achieve their optimal model utility through collaborating with $I^{2}$ and $I^{3}$. In this case, $C = \left\{I^{1}, I^{2},I^{3} \right\}$ denotes a collaboration coalition, and each client achieves its optimal utility by collaborating with other clients in $C$. If $I^{1}$ is taken out from $C$, $I^{3}$ will leave $C$ as well because it cannot gain any benefit through collaboration with others, and then $I^2$ will leave for the same reason. With this ring structure of $C$, none of the clients in $C$ can achieve its best performance without collaborating with the clients in $C$.
\subsection{Problem Setup}
\label{sec:3-2}
As each client in $\bm{I}$ aims to maximize its local model utility by forming a collaboration coalition with others, all clients in $\bm{I}$ can form several non-overlapping collaboration coalitions. In order to derive those coalitions, we propose the concept of \emph{collaboration equilibrium} (CE) as follows.

Suppose we have a set of coalitions $S  = \left\{C^{0}, C^{1},...C^{K} \right\}$ such that $\bigcup_{k=1}^KC^k=\bm{I}$ and $C^{k_1}\bigcap C^{k_2}=\emptyset$ for $\forall k_1\ne k_2$, then we say $S$ reaches CE if it satisfies the following two axioms.
\begin{axiom}[Inner Agreement]
All collaboration coalitions satisfy inner agreement, i.e.,
\begin{equation}
\forall C \in S, \ \forall C' \varsubsetneqq C, \ \exists I^{i} \in C', \ \text{s.t.,} \ U_{max}(I^{i}, C') < U_{max}(I^{i}, C)
\label{eq:ia}
\end{equation}
\label{axiom:1}
\end{axiom}
From Axiom~\ref{axiom:1}, inner agreement emphasizes that the clients of each coalition agree to form this coalition. It gives the necessary condition for a collaboration coalition to be formed such that any of the subset $C' \varsubsetneqq C$ can benefit from the collaboration with $C \backslash C'$. Eq.(\ref{eq:ia}) tells us that there always exists a client in $C'$ that opposes leaving $C$ because its utility will go down if $C'$ is split from $C$.
In this way, inner agreement guarantees that all coalitions will not fall apart or the clients involved will suffer. For example, $S = \left\{ \left\{I^{1},I^{2},I^{3},I^{4}  \right\}, \left\{ I^{5}, I^{6} \right\} \right\}$ in Figure~\ref{fig:example2} does not satisfy inner agreement, because the clients in the subset $ C' = \left\{I^{1},I^{2},I^{3} \right\}$ achieves their optimal utility in $C'$ and can leave $C = \left\{I^{1},I^{2},I^{3},I^{4}  \right\}$ without any loss.

\vspace{.5em}
\begin{axiom}[Outer Agreement]
The collaboration strategy should satisfy outer agreement, i.e.,
\begin{equation}
\forall C' \notin S, \ \exists I^{i} \in C', \ \text{s.t.} \ U_{max}(I^{i}, C') \leq U_{max}(I^{i}, C ) \ (I^{i} \in C \in S)
\label{eq:oa}
\end{equation}
\label{axiom:2}
\end{axiom}\vspace{-2em}
From Axiom~\ref{axiom:2}, outer agreement guarantees that there is no other coalition $C'$ which can benefit each client involved more than $S$ achieves. Eq.(\ref{eq:oa}) tells us that if $C'$ is a coalition not from $S$, there always exists a client $I^i$ and a coalition in $C\in S$ such that $I^i$ can benefit more. The collaboration strategy $S = \left\{ \left\{ I^{1},I^{2},I^{3}\right\}, \left\{I^{4} \right\}, \left\{ I^{5},I^{6} \right\} \right\}$ in Figure~\ref{fig:example2} is a CE in which the clients in $\left\{I^{1},I^{2},I^{3}\right\}$ and $\left\{ I^{5},I^{6} \right\}$ achieve their optimal model utility. Though $I^{4}$ does not achieve its maximum model utility in $\left\{I^{4} \right\}$, there is no other coalitions which can attract $I^{2}$ and $I^{5}$ to form a new coalition with $I^{4}$. Therefore, all clients have no better choice but agree upon this collaboration strategy.

Our goal is to obtain a collaboration strategy to achieve CE that satisfies Axiom~\ref{axiom:1} and Axiom~\ref{axiom:2}, so that all clients achieve their optimal model utilities in the collaboration coalition. In the next section, we introduce our algorithm in detail on 1) how to derive a collaboration strategy that can achieve CE from the benefit graph and 2) how to construct the benefit graph.

\section{Collaboration Equilibrium}
In this section, we will 
introduce our framework on learning to collaborate.
Firstly, we propose an iterative graph-theory-based method to achieve CE based on a given benefit graph.

\subsection{Achieving Collaboration Equilibrium Given the Benefit Graph}
\label{sec:4-1}

In theory, there are $B_{N}$ collaboration strategies for partitioning $N$ clients into several coalitions, where $B_{N}$ is the Bell number which denotes how many solutions for partitioning a set with $N$ elements~\citep{bell1934exponential}. While exhaustive trying all partitions has exponential time complexity, in this section, we propose an iterative method for deriving a collaboration strategy that achieves CE with polynomial time complexity. Specifically, at each iteration, we search for a \emph{stable coalition} which is formally defined in~Definition \ref{def:sc} below, then we remove the clients in the \emph{stable coalition} and re-build the benefit graph for the remaining clients. The iterations will continue until all clients are able to identify their own coalitions. 

\begin{definition}[Stable Coalition]
\label{def:sc}
Given a client set $\bm{I}$, a coalition $C^{s}$ is stable if it satisfies
\begin{enumerate}
  \item Each client in $C^{s}$ achieves its maximal model utility, i.e.,
  \begin{equation}
  \label{eq:sc-1}
  U_{max}(I^{i}, C^{s}) = U_{max}(I^{i}, \bm{I}) \ \forall I^{i} \in C^{s}.
  \end{equation}
  \item Any sub coalition $C'\subset C^{s}$ cannot achieve the maximal utility for all clients in $C'$, i.e., \label{property:2}
  \begin{equation}
  \label{eq:sc-2}
  \forall C' \varsubsetneqq C^{s}, \ \exists I^{i} \in C', \ \text{s.t.,} \ U_{max}(I^{i}, C') < U_{max}(I^{i}, C^{s}).
  \end{equation}
\end{enumerate}
\end{definition}
From Definition \ref{def:sc}, Eq.(\ref{eq:sc-1}) means that any client in a stable coalition cannot improve its model utility further. Eq.(\ref{eq:sc-2}) states that this coalition is \emph{stable} as any sub coalition $C'$ can benefit from $C^{s} \backslash C'$. Therefore any sub coalition has no motivation to leave $C^{s}$. Eq.(\ref{eq:sc-1}) implies that a stable coalition will not welcome any other clients to join as others will not benefit the clients in $C^{s}$ further. In Figure~\ref{fig:example2}, $\left\{ I^{1}, I^{2}, I^{3} \right\}$ and $\left\{ I^{5}, I^{6} \right\}$ are the two stable coalitions. In order to identify the stable coalitions from the benefit graph, we first introduce the concept of strongly connected component in a directed graph.

\sen{

\begin{definition}[Strongly Connected Component~\citep{tarjan1972depth}]
A subgraph $G'$ is a strongly connected component of a given directed graph $G$ if it satisfies:
1) It is strongly connected, which means that there is a path in each direction between each pair of vertices in $G'$;
2) It is maximal, which means no additional vertices from $G$ can be included in $G'$ without breaking the property of being strongly connected.
\label{def:scc}
\end{definition}

Then we derive a graph-based method to obtain the collaboration coalitions that can achieve collaboration equilibrium by identifying all stable coalitions iteratively according to Theorem~\ref{theorem:1} below.}

\begin{theorem}
(Proof in Appendix) Given a client set $\bm{I}$ and its $BG(\bm{I})$, the stable coalitions are strongly connected components of $BG(\bm{I})$.
\label{theorem:1}
\end{theorem}

With Theorem~\ref{theorem:1}, we need to identify all strongly connected components of $BG(\bm{I})$, which can be achieved using the \textbf{Tarjan} algorithm~\citep{tarjan1972depth} with time complexity $O(V+E)$, where $V$ is the number of nodes and $E$ is the number of edges. Then following Eq.(\ref{eq:sc-1}), we judge whether a strongly connected component is a stable coalition by checking whether all clients have achieved their maximal model utility.
A stable coalition $C^{s}$ has no interest to collaborate with other clients, so $C^{s}$ will be removed and the remaining clients $\bm{I} \backslash C^{s}$ will continue to seek collaborations until all $N$ clients find their coalitions. In this way, we can achieve a partitioning strategy, with the details shown in Algorithm~\ref{alg:CE1}.

\begin{algorithm}
\caption{Achieving collaboration equilibrium}
\label{alg:CE1}
\KwIn{$N$ institutions $\bm{I} = \{ I^{i} \}_{i = 1}^{N}$ seeking collaborating with others}
Set original client set $C \gets \bm{I}$;\\
Set collaboration strategy $S \gets \emptyset$;\\
\While {$C \not = \emptyset $}{

  \ForAll{client $I^{i} \in C$}{
    Determine the OCS of $I^{i}$ by SPO; \\
    (detailed description in Sec~\ref{sec:4-2}) \\
  }
  Construct the benefit graph $BG(C)$; \\
  Search for all strongly connected components $\left\{ C^{1}, C^{2},...C^{k} \right\}$ of $BG(C)$ using \textbf{Tarjan} algorithm;\\
  \ForAll{i = 1, 2, 3,... k}{
    \textcolor{text}{\If{$C^{i}$ is stable coalition}{
    $C \gets C \backslash C^{i}$ ; \\
    $S \gets S \cup \{ C^{i} \}$;
  }
  }
  }
}
\KwOut{collaboration strategy $S$}
\end{algorithm}

One may wonder whether our algorithm could get stuck because none of the strongly connected components are stable coalitions as highlighted in red in Algorithm~\ref{alg:CE1}. We answer this question by giving the following proposition:

\begin{proposition}
(Proof in Appendix) In any iteration, our algorithm ensures that at least one stable coalition is identified, so it will not get stuck because none of the strongly connected components are stable coalitions.
\label{proposition:1}
\end{proposition}

The clients in all stable coalitions found in each iteration cannot improve their model utility further and will not collaborate with others because there are no additional benefits. Therefore, the collaboration strategy is approved by all clients and we have the following theorem:

\begin{theorem}
(Proof in Appendix) The collaboration strategy obtained above achieves collaboration equilibrium.
\label{theorem:3}
\end{theorem}





\begin{figure*}[h!]
    \setlength{\abovecaptionskip}{-0.1cm}
    \setlength{\belowcaptionskip}{-0.1cm}
    \centering{
    \includegraphics[width=1.9\columnwidth]{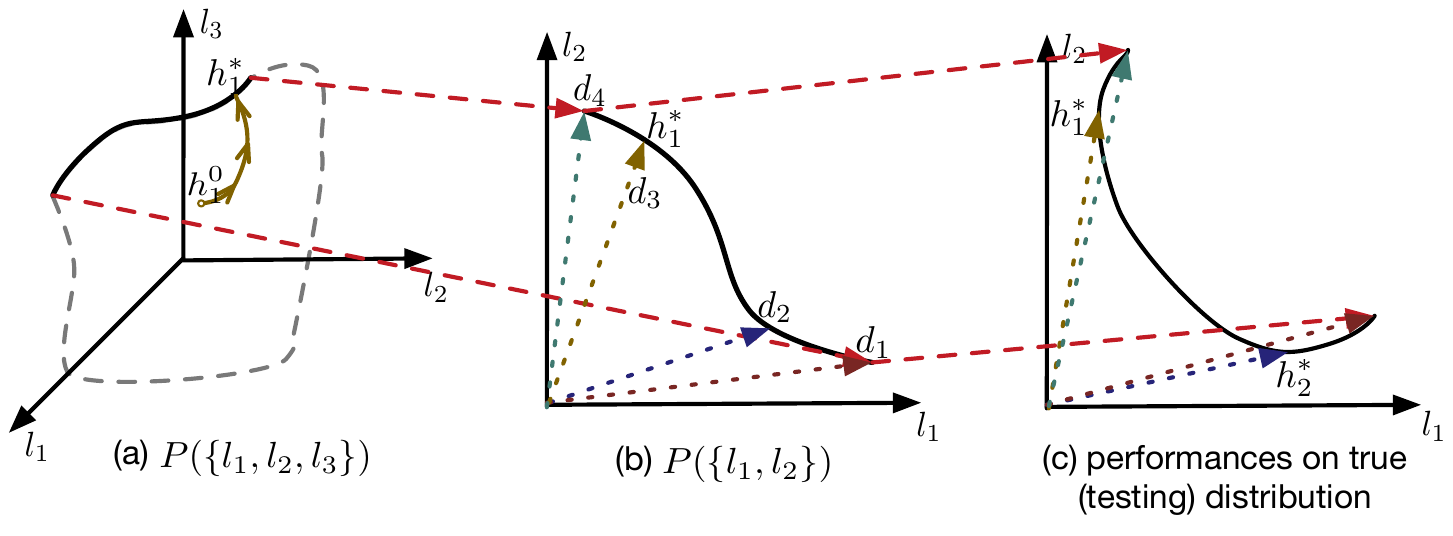}}
    \caption{(a) the loss plane of $P(\left\{l_{1}, l_{2}, l_{3} \right\})$ learned from training data of $\left\{I^{1}, I^{2}, I^{3}\right\}$; (b) the loss curve of $P(\left\{l_{1}, l_{2} \right\})$ learned from training data of $\left\{I^{1}, I^{2}\right\}$ is embedded in the loss plane of $P(\left\{l_{1}, l_{2}, l_{3} \right\})$; $d_{1},d_{2},.,d_{4}$ are 4 vectors corresponding to 4 Pareto models in $P(\left\{l_{1}, l_{2} \right\})$; (3) the performances of the models $h \in P(\left\{l_{1}, l_{2} \right\})$ on true (testing) distributions and $h_{1}^{*}$ ($h_{2}^{*}$) achieves the optimal performance on client $I^{1}$ ($I^{2}$) corresponding to $d_{3}$ ($d_{2}$) in (b).}
    \label{fig:spo}
\end{figure*}

\subsection{Determine the Benefit Graph by Specific Pareto Optimization}
\label{sec:4-2}

\cui{The benefit graph of $N$ clients consists of the clients and their corresponding OCS. However, each client has $2^{N-1}$ collaborator sets and it's hard to determine which one is the OCS. Exhaustive trying all sets to determine the OCS may be impractical, especially when there are many clients. To identify the OCS effectively and efficiently, we propose Specific Pareto Optimization (SPO) to \liang{alternately perform the following two steps:} 1). learn a Pareto model (defined below) given the weight of all clients $\bm{d}$; 2). optimize the weight vector $\bm{d}$ to search for a best model $M^{*}(I^{i})$ by gradient descent.}

\begin{definition}[Pareto Solution and Pareto Front]
We consider $n$ objectives corresponding to $n$ clients: $ l_{i} : \mathbb{R}^{n} \rightarrow \mathbb{R}_{+}, i = \left\{ 1,2,...,n \right\}$. Given a learned hypothesis $h$, suppose the loss vector $\bm{l(h)} = [l_{1}, l_{2},...,l_{n} ]$ represents the utility loss on $n$ clients with hypothesis $h \in \mathcal{H}$, we say $h$ is a Pareto Solution if there is no hypothesis $h'$ that dominates h, often called Pareto optimality, i.e., $$\nexists h^{\prime} \in \mathcal{H}, \text { s.t. } \forall i: l_{i}\left(h^{\prime}\right) \leq l_{i}(h) \text { and } \exists j: l_{j}\left(h^{\prime}\right)<l_{j}(h).$$ In a collaboration network with $N$ clients $\{ I^{i}\}_{i=1}^{N}$, as each client has its own learning task which can be formulated as a specific objective, we use $P( \left\{ I^{1}, I^{2},...,I^{N} \right\} )$ to represent the Pareto Front (PF) of the client set $\{ I^{i}\}_{i=1}^{N}$ formed by all Pareto hypothesis.
\label{def:pt}
\end{definition}

\paragraph{Learning a best model and identify the OCS by SPO} \cui{ To search for an optimal model $M^{*}(I^{i})$ for $I^{i}$, we propose to firstly learn the empirical Pareto Front by a hypernetwork denoted as $HN$ \footnote{More information please refer to~\citep{navon2020learning}}. As each client owns its objective, given the weight of all clients (objectives), the learned PT by $HN$ outputs a Pareto model $M$,
\begin{equation}
\setlength{\abovedisplayskip}{4pt}
    M = HN(\bm{d}),
\end{equation}
where $\bm{d} = [d^{1}, d^{2},...,d^{N}]$ satisfying $\sum_{i=1}^{N} d^{i} = 1$ and $d^{i}$ denotes the weight of the objective $l_{i}$. Each $\bm{d}$ corresponds to a specific Pareto model $M$, and all Pareto models $M$ satisfy Pareto optimality that the losses on the training data of all clients cannot be further optimized.

Though the Pareto models maximize the utilization of the training data from all clients, they may not be the best models on the true (test) data distribution. For example, $HN(\bm{d}_{4})$ achieves a minimum loss on the training data of $I^{1}$ shown in Figure~\ref{fig:spo} (b), but it is not the optimal model on the true data distribution of $I^{1}$ shown in Figure~\ref{fig:spo} (c). Therefore, we propose to search for a best Pareto model that achieves the best performance on the true (validation) data set of the target client $I^{i}$, i.e.,
\begin{equation}
\setlength{\belowdisplayskip}{-5pt}
    M^{*}(I^{i}) = HN(\bm{d}^{*}), \quad \text{where} \  \bm{d}^{*} = \arg\max_{\bm{d}} \mathrm{Per}(HN(\bm{d}), I^{i}),
\label{eq:hn_per}
\end{equation}

where $\mathrm{Per}(HN(\bm{d}), I^{i})$ denotes the performance of $HN(\bm{d})$ on the validation data set of $I^{i}$. We identify the OCS of $I^{i}$ based on the optimized weight of all clients $\bm{d^{*}}$. For example shown in Figure~\ref{fig:spo}(c), the best model is $h_{1}^{*}$ for client $I^{1}$ and the corresponding optimized weight $\bm{d}^{*} = \bm{d}_{3} = [0.7, 0.3, 0]$. So the OCS of $I^{1}$ consists of the clients with non-zero weights, which is $\left\{ I^{1}, I^{2}\right\}$. }

\begin{proposition}[Pareto Front Embedding Property]
\label{prop:embed-pt}
(proof in Appendix) Suppose $\bm{l'^{*}} = [l_{i}(h'^{*})], i \in C'$ and $\bm{l^{*}} = [l_{i}(h^{*})], i \in C$ are the loss vectors achieved by the PFs $P(C')$ and $P(C)$ where $C' \subset C$, then
\begin{equation}
\forall h'^{*} \in P(C'), \ \exists h^{*} \in P(C), \ \text{s.t.,} \ l_{i}(h'^{*}) = l_{i}(h^{*}) \ \forall i \in C'.
\label{eq:prop}
\end{equation}
\end{proposition}
From Proposition \ref{prop:embed-pt}, the loss vectors achieved by the PF of a sub-coalition are embedded in the loss vectors of the full coalition, such as the loss curve of $P(\left\{ l_{1}, l_{2} \right\})$ is in the loss plane of $P(\left\{ l_{1}, l_{2}, l_{3} \right\})$ shown in Figure~\ref{fig:spo} (a) and (b).
\paragraph{Explanation of SPO from a geometric point of view} \cui{
Intuitively, if the best model $M^{*}(I^{i})$ searched on the PF of all clients $P(\bm{I})$ belongs to the PF of a sub coalition $P(C)$ simultaneously, i.e.,
\begin{equation}
    \setlength{\abovecaptionskip}{-5pt}
    \setlength{\belowcaptionskip}{-5pt}
    M^{*}(I^{i}) \in P(C) \quad \text{and} \quad M^{*}(I^{i}) \in P(I),
\end{equation}
then the clients $\bm{I} \backslash C$ are not necessary for obtaining $M^{*}(I^{i})$. However, exhaustively trying the PF of all sub-coalitions can have exponential time complexity. From Proposition~\ref{prop:embed-pt}, the PF of full coalition $\bm{I}$ contains the PF of all sub-coalitions. Therefore, we propose SPO to search for a best model $M^{*}(I^{i})$ on the PF of full coalition $\bm{I}$, and identify whether $M^{*}(I^{i})$ belongs to the PF of a sub-coalition using the optimized weight vector $\bm{d}$. For example, suppose there are three clients and the three corresponding objectives achieved by the PF are shown in Figure~\ref{fig:spo}(a). The model $h^{*}_{1}$ in Figure~\ref{fig:spo}(a) is on the PF $P(\left\{ l_{1}, l_{2}, l_{3} \right\})$. Since the corresponding weight $\bm{d} = [0.7, 0.3, 0]$, from Figure~\ref{fig:spo}(b), $h^{*}_{1}$ is also on the PF $P(\left\{ l_{1}, l_{2} \right\})$, so $I^{3}$ is not a necessary client and the OCS of $I^{1}$ is $\left\{ l_{1}, l_{2} \right\}$.
}

\subsection{More Discussion}
\textbf{Time Complexity Analysis} The analysis about the time complexity is as follows:
\begin{enumerate}
    \item to find an OCS efficiently, we propose SPO to learn an optimal model by gradient descent and identify the OCS based upon the geometric location of the learned model on the Pareto Front according to Proposition 1, which avoids exhaustively trying;
    \item to find the stable coalitions of a given client set, we propose to identify the OCS of all clients to construct the benefit graph firstly. Then we propose a graph-based method to recognize the stable coalitions which has $O(V+E)$ time complexity as stated in Sec~\ref{sec:4-1};
    \item to achieve a collaboration equilibrium, we propose to look for stable coalitions and remove them iteratively. Combining (1) and (2), our method to reach a CE has polynomial time complexity.
\end{enumerate}

\noindent \textbf{Influence Factor of CE} The collaboration equilibrium of a collaborated network is affected by 1) the selected model structure; 2) the data distribution of the clients. For 1), different model structures utilize the data in different ways. For example, non-linear models can capture the non-linear mapping relations in the data among different local clients while linear models cannot. For 2), the collaboration equilibrium is affected by the data distribution of the clients given the model structure. The data distribution of one client determines whether it can benefit another when learning together and this was verified by the experiments on the synthetic dataset.

\section{Experiments}
To demonstrate the effectiveness of SPO, we conduct experiments on synthetic data, a real-world UCI dataset \textbf{Adult}~\citep{kohavi1996scaling} and a benchmark data set \textbf{CIFAR10}~\citep{lecun1998gradient}. We use the following two ways to demonstrate the effectiveness of our proposed SPO:

\begin{enumerate}
    \item exhaustive trying; we exhaustively try each subset to obtain the true OCS for each client. Then we compare the true benefit graph with the learned benefit graph by SPO;
    \item performance of the learned models; we compare the model performance learned by SPO with prior methods in PFL, as the optimality of the obtained benefit graph is reflected in the performance of the model learned based on the benefit graph.
\end{enumerate}

To intuitively show the motivation of \emph{collaboration equilibrium} and the practicability of our framework, we conduct experiments on a real-world multiple hospitals collaboration network using the electronic health record (EHR) data set \textbf{eICU}~\citep{pollard2018eicu}.

As SPO aims to achieve an optimal model utility by optimizing the personalized model on the PF of all clients, we use SPO to denote the model utility achieved by SPO. According to the OCS determined by SPO, we achieve a CE for all clients and the model utility of each client in the CE can be different from the utility achieved by SPO. We use CE to denote the model utility achieved in the CE without causing further confusion. We anonymously upload our source code in Supplementary Material. More implementation details can be found in Appendix.

\subsection{Synthetic Experiments}
\noindent \textbf{Synthetic data} Suppose there are 6 clients in the collaboration network. The synthetic features owned by each client $I^{i}$ are generated by $\mathbf{x} \sim \mathcal{U}[-1.0,1.0]$; the ground-truth weights $\mathbf{u}_{i}=\mathbf{v}+\mathbf{r}_{i}$ are samples as $\mathbf{v} \sim \mathcal{U}[0.0,1.0], \mathbf{r}_{i} \sim \mathcal{N}(\mathbf{0}_{d}, \rho^{2})$  where $\rho^{2}$ represents the client variance (if $\rho^{2}$ increases, the data distribution discrepancy among clients will increase). Labels of the clients are observed with i.i.d noise $\epsilon \sim \mathcal{N}\left(0, \sigma^{2}\right)$. To generate conflicting learning tasks assigned to different clients, we flip over the label of some clients: $y = \mathbf{u}_{i}^{\top} \mathbf{x}+\epsilon, i \in \left\{0,1,2\right\}$ and $y = -\mathbf{u}_{i}^{\top} \mathbf{x}+\epsilon, i \in \left\{3,4,5\right\}$.
\begin{figure}[h!]
    \centering{
    \includegraphics[width=0.8\columnwidth]{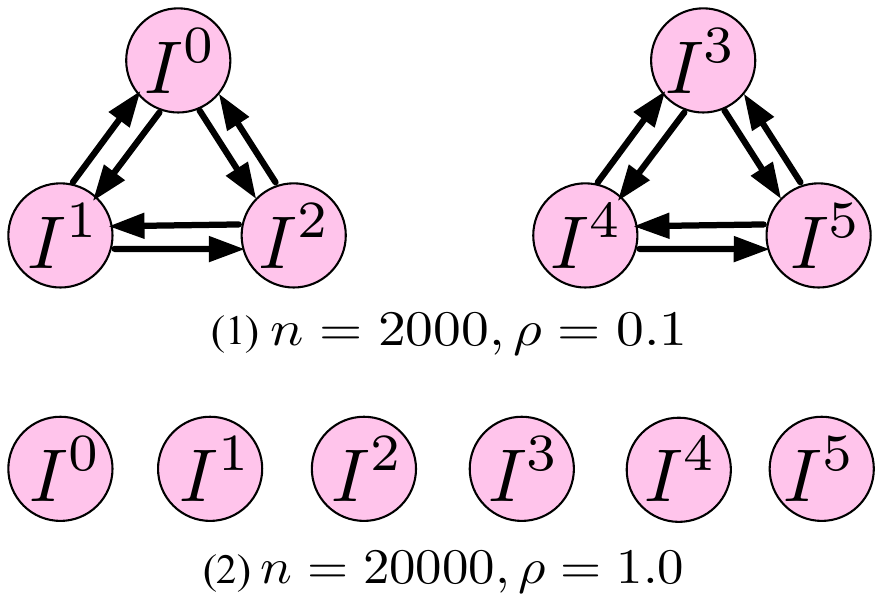}}
    \caption{The true benefit graph by exhaustive trying}
    \label{fig:syn_bg}
\end{figure}

To obtain the true benefit graph, we exhaustively try all subsets to determine the OCS for each client. For example, we learn $2^{5}$ models for each group of clients for $I^{0}$ and determine the OCS with which the model achieves the minimal loss. The true benefit graph is shown in Figure~\ref{fig:syn_bg}.
\begin{figure*}[h!]
\setlength{\abovecaptionskip}{-3pt}
\setlength{\belowcaptionskip}{-3pt}
    \centering{
    \includegraphics[width=1.8\columnwidth]{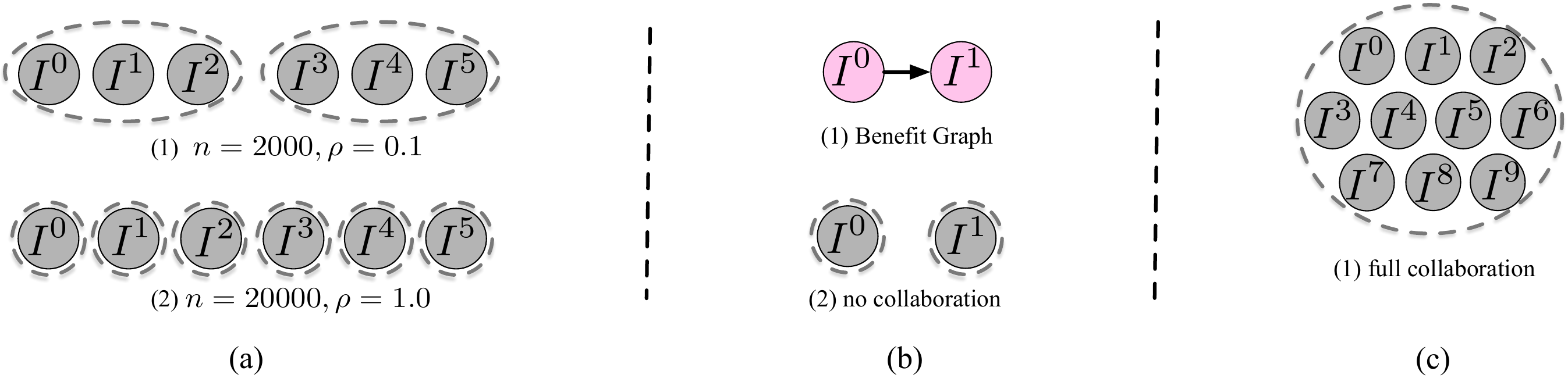}}
    \caption{Collaboration equilibrium on synthetic data, Adult and CIFAR10.}
    \label{fig:syn-equ}
\end{figure*}

\begin{table}[htbp]
\setlength{\abovecaptionskip}{-0pt}
\centering
\caption{synthetic}
\begin{tabular}{lllll}
\toprule
\multirow{2}{*}{I} & \multicolumn{2}{c}{$n = 2000, \rho = 0.1$} & \multicolumn{2}{c}{$n = 20000, \rho = 1.0$} \\
\cline{2-3}
\cline{4-5}
        & OCS                                  & CE (MSE)   & OCS                    &  CE (MSE)  \\
\hline
$I^{0}$ &$\left\{I^{0}, I^{1}, I^{2} \right\}$ & 0.24±0.08 & $\left\{I^{0}\right\}$ & 1e-4±.0\\
$I^{1}$ &$\left\{I^{0}, I^{1}, I^{2} \right\}$ & 0.26±0.08 & $\left\{I^{1}\right\}$ & 1e-4±.0\\
$I^{2}$ &$\left\{I^{0}, I^{1}, I^{2} \right\}$ & 0.24±0.04 & $\left\{I^{2}\right\}$ & 1e-4±.0\\
$I^{3}$ &$\left\{I^{3}, I^{4}, I^{5} \right\}$ & 0.26±0.07 & $\left\{I^{3}\right\}$ & 1e-4±.0\\
$I^{4}$ &$\left\{I^{3}, I^{4}, I^{5} \right\}$ & 0.26±0.09 & $\left\{I^{4}\right\}$ & 1e-4±.0\\
$I^{5}$ &$\left\{I^{3}, I^{4}, I^{5} \right\}$ & 0.26±0.03 & $\left\{I^{5}\right\}$ & 1e-4±.0\\
\bottomrule
\end{tabular}
\label{table:synthetic}
\end{table}

We also show the experimental results of SPO in Table~\ref{table:synthetic}. From Table~\ref{table:synthetic}, when there are fewer samples ($n=2000$) and less distribution discrepancy $\rho = 0.1$ in the client set $\left\{I^{0}, I^{1}, I^{2}\right\}$ and $\left\{I^{3}, I^{4}, I^{5}\right\}$ with similar label generation process, these clients collaborate with others to achieve a low MSE. In this case, the OCS of each client is the clients with similar learning tasks and we achieve CE $S = \{ \{I^{i} \}_{i=0}^{2}, \{I^{i} \}_{i=3}^{5} \}$ as shown in the top of Figure~\ref{fig:syn-equ} (a). With the increase of the number of samples and the distribution discrepancy, collaboration cannot benefit the clients and all clients will learn individually on their own data. Therefore, when $n = 20000$ and $\rho = 1.0$, the OCS of each client is itself and the collaboration strategy $S = \{ \{I^{0} \}, \{I^{1} \},., \{I^{5}\} \}$ leads to a CE as shown in the bottom of Figure~\ref{fig:syn-equ} (a).

Comparing the OCS in Table~\ref{table:synthetic} with the graph in Figure~\ref{fig:syn_bg}, SPO obtains the same benefit graph realized by exhaustively trying in polynomial time complexity.

\begin{table}[!htp]
\setlength{\abovecaptionskip}{-1pt}
\begin{minipage}[t]{0.26\textwidth}
\makeatletter\def\@captype{table}
\caption{Adult}
\begin{spacing}{1.815}
\begin{tabular}{ccc}
\toprule
\multirow{2}{*}{methods} & \multicolumn{2}{c}{Accuracy}  \\
\cline{2-3}
           & $I^{0}$    & $I^{1}$    \\
\hline
AFL        & 82.6 ± .5  & 73.0 ± 2.2 \\
q-FFL      & 82.4 ± .1  & 74.4 ± .9 \\
local      & \textbf{83.5 ± .0}  & 66.9 ± 1.0\\
SPO (ours)  & 82.8 ± .3  & \textbf{77.0 ± .7} \\
\hline
CE (ours)       & \textbf{83.5 ± .0}  & 66.9 ± 1.0\\
\bottomrule
\end{tabular}
\end{spacing}
\label{table:adult}
\end{minipage}
\begin{minipage}[t]{0.21\textwidth}
\makeatletter\def\@captype{table}
\caption{CIFAR10}
\begin{spacing}{1.27}
\begin{tabular}{cc}
\toprule
\multicolumn{1}{c}{\textbf{methods}} & \textbf{accuracy} \\
\hline
Local                                & 86.46 ± 4.02      \\
FedAve                               & 51.42 ± 2.41      \\
Per-FedAve                           & 76.65 ± 4.84      \\
FedPer                               & 87.27 ± 1.39      \\
pFedMe                               & 87.69 ± 1.93      \\
LG-FedAve                            & 89.11 ± 2.66      \\
pFedHN                               & 90.83 ± 1.56      \\
SPO (ours)                           & \textbf{92.47 ± 4.80}      \\
\hline
CE  (ours)                                 & \textbf{92.47 ± 4.80}      \\
\bottomrule
\end{tabular}
\end{spacing}
\label{table:cifar10}
\end{minipage}
\end{table}
\noindent \textbf{UCI adult data} UCI \emph{adult} is a public dataset~\citep{kohavi1996scaling}, which contains more than 40000 adult records and the task is to predict whether an individual earns more than 50K/year given other features (e.g., age, gender, education, etc.). Following the setting in~\citep{li2019fair,mohri2019agnostic}, we split the data set into two clients. One is PhD client ($I^{1}$) in which all individuals are PhDs and the other is non-PhD client ($I^{0}$). We implement SPO on this data set. Specifically, we construct the hypernetwork using a 1-layer hidden MLP. The target network is a Logistic Regression model as in ~\citep{li2019fair,mohri2019agnostic}. We split 83\% training data for learning the PF and the remaining 17\% training data for determining the optimal vector $\bm{d}$. We compare the performance with existing relevant methods AFL \citep{mohri2019agnostic} and q-FFL \citep{li2019fair}\footnote{The results of baselines are from~\citep{li2019fair}}.

The two clients $I^{0}$ and $I^{1}$ have different data distribution and non-PhD client has more than 30000 samples while PhD client has about 500 samples. From Table \ref{table:adult}, PhD client improves its performance by collaborating with non-PhD client. However, non-PhD client achieves an optimal accuracy (83.5) by local training, so non-PhD will not be willing to collaborate with non-PhD client. Therefore, the benefit graph is shown at the top of Figure \ref{fig:syn-equ} (b). The CE is non-collaboration as in the bottom of Figure \ref{fig:syn-equ} (b) and the model of both clients in the CE are trained individually.

Compared with prior methods, SPO achieves higher performance on both clients in Table~\ref{table:adult} especially on PhD client (77.0), which verifies its superiority on the learning of personalized models.

\begin{figure*}[h!]
  \centering
  \subfigure[benefit graph]{
    \centering
    \includegraphics[width=0.5\columnwidth]{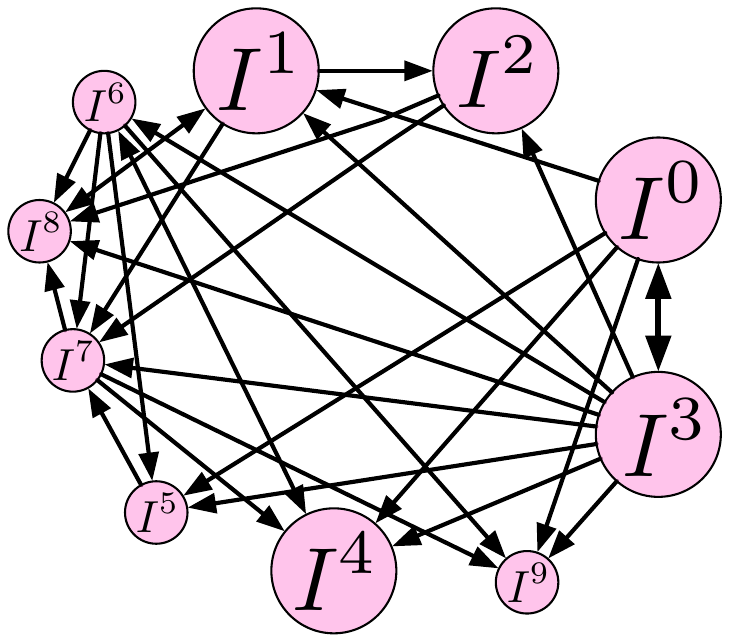}
    \label{fig:eicu_bg}
  }%
  \hskip 0.5in
  \subfigure[strongly connected components of (a)]{
    \centering
    \includegraphics[width=0.5\columnwidth]{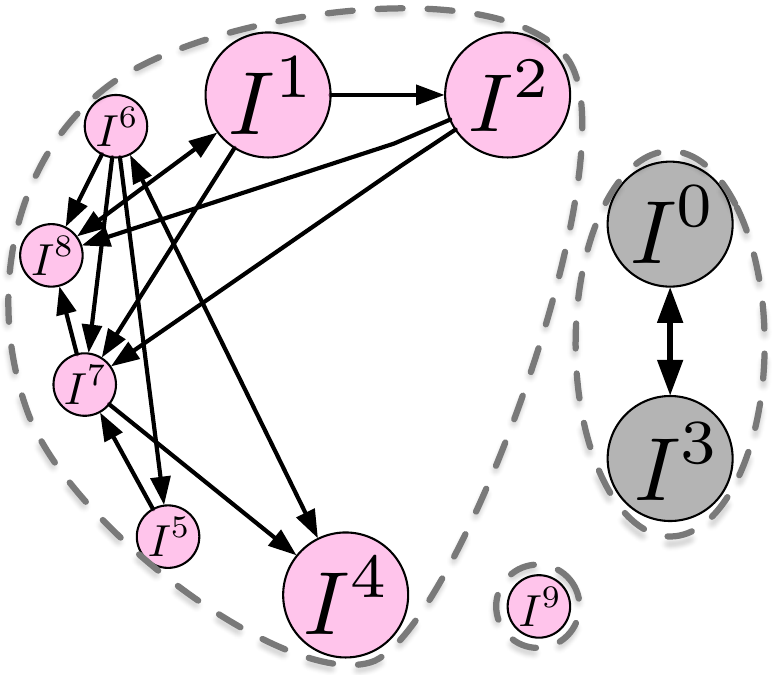}
    \label{fig:eicu_scc}
  }%
  \hskip 0.5in
  \subfigure[collaboration equilibrium]{
    \centering
    \includegraphics[width=0.5\columnwidth]{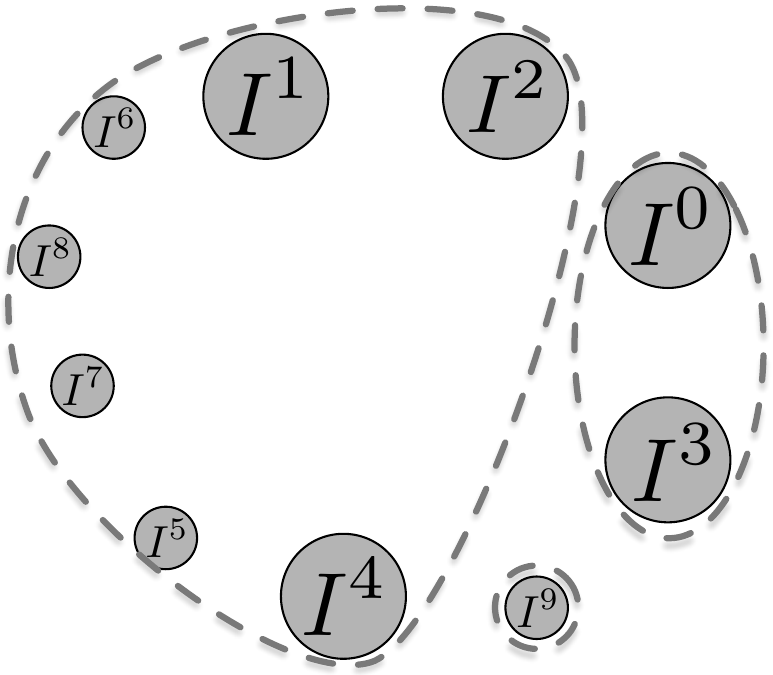}
    \label{fig:eicu_ce}
  }%
  \caption{Collaboration Equilibrium of 10 real hospitals}
  \label{fig:eicu}
\end{figure*}
\begin{table*}[htbp]
\setlength{\abovecaptionskip}{-0pt}
\centering
\caption{eICU}
\begin{tabular}{ccccccccccc}
\toprule
\multirow{2}{*}{methods} & \multicolumn{10}{c}{AUC}  \\
\cline{2-11} \\
       &$I^{0}$ &$I^{1}$ &$I^{2}$ &$I^{3}$ &$I^{4}$ &$I^{5}$ &$I^{6}$ &$I^{7}$ &$I^{8}$ &$I^{9}$  \\
\hline
Local     &66.89 &85.03 &61.83 &68.83 &82.31 &59.65 &67.78 &40.00 &61.90 &70.00  \\
FedAve    &71.92 &89.36 &\textbf{81.00} &\textbf{73.89} &80.23 &70.18 &52.22 &40.00 &61.90 &75.00  \\
SPO (ours) &\textbf{76.35} &\textbf{91.80} &80.28 &70.52 &\textbf{86.93} &\textbf{82.46} &\textbf{71.11} &\textbf{40.00} &\textbf{76.19} &\textbf{83.33}  \\
\hline
CE (ours)      &77.93 &87.28 &70.47 &70.64 & 83.48 &64.92 & 68.89 & 45.00 & 61.90 & 70.00 \\
\bottomrule
\end{tabular}
\label{table:eicu}
\end{table*}

\subsection{Benchmark Experiments}
We compare our method with previous personalized federated learning (PFL) methods on CIFAR-10~\citep{krizhevsky2009learning}\footnote{The results of baselines are from ~\citep{shamsian2021personalized}}. CIFAR10 is a public dataset~\citep{lecun1998gradient} which contains 50000 images for training and 10000 images for testing. In our experiments, we follow the setting in ~\citep{mcmahan2016federated}. We split the data into 10 clients and simulate a non-i.i.d environment by randomly assigning two classes to each client among ten total classes. The training data of each client will be divided into a training set (83\%) and a validation set(17\%). We construct the hypernetwork using 3-layer hidden MLP for generating the parameters of the target network and the target network is constructed following the work~\citep{shamsian2021personalized}. All baselines share the same target network structure for each client.

Baselines we evaluate are as follows: (1) Local training on each client; (2) FedAvg~\citep{mcmahan2016federated}; (3) Per-FedAvg~\citep{fallah2020personalized}, a meta-learning based PFL algorithm. (4) pFedMe~\citep{t2020personalized}, a PFL approach which adds a Moreau-envelopes loss term; (5) LG-FedAvg~\citep{liang2020think} PFL method with local feature extractor and global output layers; (6) FedPer~\citep{arivazhagan2019federated}, a PFL approach that learns personal classifier on top of a shared feature extractor; (7) pFedHN~\citep{shamsian2021personalized}, a PFL approach that generates models by training a hyper-network. In all experiments, our target network shares the same architecture as the baseline models. For each client, we split 87\% of the training data for learning a Pareto Front by collaborating with the others and the remaining 13\% of the training data for optimizing the vector $\bm{d}$ to reach an optimal model as shown in Figure~\ref{fig:spo} (c).

Table~\ref{table:cifar10} reports the results of all methods. FedAve achieves a lower accuracy (51.4) compared to local training (86.46) which means that training a global model can hurt the performance of each client. Compared to other PFL methods in Table~\ref{table:cifar10}, SPO reaches an optimal model on the PF of all objectives and achieves a higher accuracy (92.47). As the features learned from the images are transferable though there is a label shift among all clients, the collaboration among all clients leads to a more efficient feature extractor for each client. Therefore, the benefit graph of this collaboration network is a fully connected graph and the collaboration equilibrium is that all clients form a full coalition for collaboration as shown in Figure~\ref{fig:syn-equ} (c). In this experiment, the accuracy model of each client in CE equals to the accuracy achieved by SPO.

\subsection{Hospital Collaboration}
eICU~\citep{pollard2018eicu} is a clinical data set collecting the patients about their admissions to ICUs with hospital information. Each instance is a specific ICU stay. We follow the data pre-processing procedure in~\citep{sheikhalishahi2019benchmarking} and naturally treat different hospitals as local clients. We conduct the task of predicting in-hospital mortality which is defined as the patient’s outcome at the hospital discharge. This is a binary classification task, where each data sample spans a 1-hour window. In this experiment, we select 5 hospitals with more patient samples (about 1000) $\left\{ I^{i} \right\}_{i=0}^{4}$ and 5 hospitals with less patient samples $\left\{ I^{i} \right\}_{i=5}^{9}$ (about 100). Due to label imbalance (more than 90\% samples have negative labels), we use AUC to measure the utility for each client as in~\citep{sheikhalishahi2019benchmarking}. We construct the hypernetwork by a 1-layer MLP for training the Pareto Front of all objectives. The target network is an ANN model following the work in ~\citep{sheikhalishahi2019benchmarking}. For a fair comparison, we also use the same ANN in~\citep{sheikhalishahi2019benchmarking} all baselines.

The model AUC of each hospital is reported in Table~\ref{table:eicu}. Because of the lack of patient data for each hospital, Local achieves a relatively lower AUC compared to FedAve and SPO. For example, due to the severely insufficient data, the AUC of $I^{7}$ is very low (0.4), which indicates that there is a drastic shift between test distribution and training distribution. While patient populations vary substantially from hospital to hospital, SPO learns a personalized model for each hospital and outperforms FedAve from Table~\ref{table:eicu}.

\paragraph{Collaboration Equilibrium}
\cui{The benefit graph of all hospitals determined by SPO and its corresponding strongly connected components are shown in Figure~\ref{fig:eicu_bg} and~\ref{fig:eicu_scc}. From Figure~\ref{fig:eicu_bg}, $I^{0}$ and $I^{3}$ are the unique necessary collaborator for each other, $C^1 = \left\{I^{0}, I^{3} \right\}$ is the first identified stable coalition as shown in Figure~\ref{fig:eicu_scc}; $I^{9}$ is a tiny clinic that cannot contribute to any hospitals, so no hospital is willing to collaborate with it and $I^{9}$ learns a local model with its own data by forming a simple coalition $C^{2} = \left\{I^{9} \right\}$. The benefit graph of the remaining clients are shown in Figure~\ref{fig:eicu_scc}. On the one hand they cannot benefit $I^{3}$ or $I^{0}$ so they cannot form coalitions with them, on the other hand they refuse to contribute $I^{9}$ without any charge. They choose form the coalition $C^{3} = \left\{I^{1}, I^{2}, I^{4}, I^{5}, I^{6}, I^{7}, I^{8} \right\}$ to maximize their AUC. Therefore, the CE in this hospital collaboration network is achieved by the collaboration strategy $S = \{C^{1}, C^{2}, C^{3} \}$ and the model AUC of each client in the CE is in Table~\ref{table:eicu}.}

\section{Applications}
One cannot know which clients should collaborate with unless it knows the result of the collaboration. This requires that all clients agree to collaborate to obtain the results of the collaboration, before finalizing the collaboration equilibrium. The premise of achieving the collaboration equilibrium is that all clients agree to collaborate first to construct the benefit graph. In practice, this would be done by an impartial and authoritative third-party (e.g., the industry association) in the paradigm of federated learning. The approved third-party firstly determines the benefit graphs by learning the optimal models from multiple clients without direct access to the data of the local clients. Then the approved third-party derives the collaboration equilibrium based upon the benefit graphs and publishes the collaboration strategies of all clients.

As our framework quantifies the benefit and the contribution of each client in a collaborative network, on the one hand, this promotes a more equitable collaboration such as some big clients may no longer collaborate with small clients without any charge; on the other hand, it also leads to a more efficient collaboration as all clients will collaborate with the necessary collaborators rather than all participants.
\begin{acks}
Sen Cui, Weishen Pan and Changshui Zhang would like to acknowledge the funding by the National Key Research and Development Program of China (No. 2018AAA0100701). Kun Chen would like to acknowledge the support from the U.S. National Institutes of Health (R01-MH124740).
\end{acks}
\section{Conclusion}
In this paper, we propose a \emph{learning to collaborate} framework to achieve collaboration equilibrium such that any of the individual clients cannot improve their performance further. Comprehensive experiments on benchmark and real-world data sets demonstrated the validity of our proposed framework. In our study, some small clients could be isolated as they cannot benefit others. Our framework can quantify both the benefit to and the contribution from each client in a network. In practice, such information can be utilized to either provide incentives or to impose charges on each client, to facilitate and enhance the foundation of the network or coalition.

\bibliographystyle{ACM-Reference-Format}
\bibliography{kdd_2022}

\clearpage
\appendix
\section{Proofs of all Theoretical Results}

\subsection{Proof of Theorem 1}
To prove Theorem~\ref{theorem:1}, we first prove that the benefit graph of a stable coalition is strongly connected shown in Lemma~\ref{lemma:1}.

\begin{lemma}
For a given client set $\bm{I}$, the benefit graph of each stable coalition $C^{s}$ is strongly connected, which means that there is a path in each direction between each pair of vertices in $BG(C^{s})$.
\label{lemma:1}
\end{lemma}

\begin{proof}
\label{proof:lemma1}
We will prove the Lemma~\ref{lemma:1} by contradiction. Given a stable coalition $C^{s}$, suppose there exsits a pair of vertices $I^{1}, I^{2} \in BG(C^{s})$ such that there is no path from $I^{1}$ to $I^{2}$, which is denoted as $\nexists p, \ \text{s.t.,} \ I^{1} \rightarrow I^{2}$ for expressive clearly. We split $C^{s}$ into two sub-coalitions $C'$ and $C^{s} \backslash C'$ depending on whether the clients have paths to $I^{2}$. The clients in $C'$ have no path to $I^{2}$, and the clients in $C^{s} \backslash C'$ have paths to $I^{2}$.
\begin{equation}
\begin{aligned}
& \forall I^{i} \in C', \nexists p, \ \text{s.t.,} \ I^{i} \rightarrow I^{2} \\
& \forall I^{i} \in C^{s} \backslash C', \exists p, \ \text{s.t.,} \ I^{i} \rightarrow I^{2}. \\
\end{aligned}
\end{equation}
Obviously, $C'$ and $C^{s} \backslash C'$ are not empty because $I^{1} \in C'$ and and $I^{2} \in C^{s} \backslash C'$. From Eq.(5) in the main text in the Definition of \emph{stable coalition}, each client $I^{i} \in C^{s}$ achieves its maximal utility by collaborating with others in $C^{s}$ which means that

\begin{equation}
\begin{aligned}
\forall I^{i} \in C^{s}, \ \forall I^{j} \in C^{opt}_{\bm{I}}(I^{i}), \ I^{j} \in C^{s}.
\end{aligned}
\end{equation}

From the Definition of \emph{Optimal Collaborator Set} (OCS), each client $I^{i}$ in the OCS of $I^{2}$ has an edge from $I^{i}$ to $I^{2}$,
\begin{equation}
\begin{aligned}
& \forall I^{i} \in C^{opt}_{\bm{I}}(I^{2}), \ \exists p ,\ \text{s.t.,} \ I^{i} \rightarrow I^{2}, \\
\end{aligned}
\end{equation}

Therefore, because each client $I^{i} \in C^{opt}_{\bm{I}}(I^{2})$ is in $C^{s} \backslash C'$, $I^{2}$ can achieve its maximal utility in $C^{s} \backslash C'$. Morever, we can prove that $I^{i} \in C^{opt}_{\bm{I}}(I^{2})$ achieves its maximal utiltiy in $C^{s} \backslash C'$ because each client $I^{i} \in C^{opt}_{\bm{I}}(I^{2})$ and its corresponding OCS are in $C^{s} \backslash C'$,

\begin{subequations}
\begin{align}
& \forall I^{i} \in C^{opt}_{\bm{I}}(I^{2}), \ \forall I^{j} \in C^{opt}_{\bm{I}}(I^{i}), \ \exists p, \ I^{j} \rightarrow I^{i}, \label{eq:4-1}\\
& \forall I^{i} \in C^{opt}_{\bm{I}}(I^{2}), \ \exists p ,\ \text{s.t.,} \ I^{i} \rightarrow I^{2}, \label{eq:4-2}\\
& \Rightarrow \ \forall I^{i} \in C^{opt}_{\bm{I}}(I^{2}), \ \forall I^{j} \in C^{opt}_{\bm{I}}(I^{i}), \ \exists p, \ I^{j} \rightarrow  I^{i} \rightarrow I^{2}
\end{align}
\label{eq:4}
\end{subequations}

From Eq.(\ref{eq:4}), the clients in the OCS of $I^{i}$ ($ I^{i} \in C^{opt}_{\bm{I}}(I^{2})$) are in $C^{s} \backslash C'$ and $I^{i}$ can achieve its optimal utility. Based on the same analysis, for each client $I^{i} \in C^{s} \backslash C'$, its OCS $C^{opt}_{\bm{I}}(I^{i})$ is in $C^{s} \backslash C'$ because all clients in $C^{opt}_{\bm{I}}(I^{i})$ have a path to $I^{2}$. So we have the conclusion that all clients in $C^{s} \backslash C'$ achieves its optimal utility by collaborating with others in $C^{s} \backslash C'$. However, this contradicts Eq.(6) in the main text in the Definition of \emph{stable coalition}. Therefore, we prove that the benefit graph $BG(C^{s})$ of the stable coalition $C^{s}$ is strongly connected as in Lemma~\ref{lemma:1}.
\end{proof}

From Lemma~\ref{lemma:1}, the $BG(C^{s})$ is strongly connected, then we will prove that $BG(C^{s})$ is a strongly connected component of $BG(\bm{I})$ by pointing out that $BG(C^{s})$ is maximal which means that no additional vertices from $\bm{I}$ can be included in $BG(C^{s})$ without breaking the property of being strongly connected.

\begin{proof}
We will prove that $BG(C^{s})$ is maximal by contradiction. Suppose there is another client $I^{0} \in \bm{I} \backslash C^{s}$ which can be added in $BG(C^{s})$ without breaking the property of being strongly connected. Therefore, there exists $I^{i} \in C^{s}$ which has an edge from $I^{0}$ to $I^{i}$. This means that $I^{0}$ is one of the necessary collaborators of $I^{i}$ ($I^{0} \in C^{opt}_{\bm{I}}(I^{i})$) and $I^{0}$ cannot achieve its optimal utility in $C^{s}$ without collaborating with $I^{0}$. However, this contradicts Eq.(5) in the main text in the Definition of \emph{stable coalition} that all clients in $C^{s}$ can achieve its optimal utility by collaborating with others in $C^{s}$. Therefore, we prove that $BG(C^{s})$ is a strongly connected component as stated in Theorem~\ref{theorem:1}.
\end{proof}

\subsection{Proof of Proposition 1}

From Algorithm~\ref{alg:CE1}, firstly, we search for all strongly connected components (SCC) { $C^{1}, C^{2},...,C^{k}$}. From Definition~\ref{def:scc}, these SCCs consist of a partition of all clients. For convenience, we group each SCC as a point in the benefit graph.
\begin{enumerate}
    \item Suppose all SCCs are not stable coalitions, this means that each SCC $C^{i}$ cannot achieve its optimal performance without the collaboration with clients in other SCCs;

    \item from (1), for each $C^{i}$, there exists at least an edge pointing to $C^{i}$ (since it needs other SCCs' help);

    \item from graph theory, if for each node $C^{i}$, there is an edge pointing to $C^{i}$, then there exists a loop in this graph.

    \item from (3), this loop means there is a larger SCC in the benefit graph, which contradicts the definition of SCC that SCC is maximal.
\end{enumerate}
So we prove that there always exists a $C^{i}$ that no edge points to, and such a SCC is a stable coalition.

In fact, the graph of SCCs is a directed acyclic graph and there always exists a node (SCC) that no edge points to (which is called "head node"). For example in Figure 1 (a), the graph of SCCs is $\left\{ I^{1}, I^{2}, I^{3} \right\} \rightarrow \left\{ I^{4} \right\} \rightarrow  \left\{I^{5}, I^{6} \right\}$, in which $\left\{ I^{1}, I^{2}, I^{3} \right\} $ is a stable coalition that no edge points to.



\subsection{Proof of Theorem 2}
\begin{proof}
\textbf{1. All coalitions in $S = \{C^0, C^{1}, ..., C^{k} \}$ satisfies Inner Agreement as in Axiom 1.}

As each coalition $C^{i} \in S$ is a stable coalition, from Eq.(6) of the Definition of \emph{stable coalition}, we have

\begin{equation}
\forall C^{i} \in S, \ \forall C' \varsubsetneqq C^{i}, \ \exists I^{i} \in C', \ \text{s.t.,} \ U_{max}(I^{i}, C') < U_{max}(I^{i}, C^{i}).
\label{eq:ce1}
\end{equation}

Then Eq.(\ref{eq:ce1}) is the definition of \emph{Inner Agreement} in Eq.(3) in the main text. Therefore, we prove that all collaboration coalitions in $S$ satisfy inner agreement.

\textbf{2. The collaboration strategy $S$ satisfies Outer Agreement as in Axiom 2.}

We will prove it by contradiction. Suppose there exists a new nonempty coalition $C' = \{I^{i^{0}}, I^{i^{1}},..., I^{i^{k}} \} \notin S$ which satisfies

\begin{equation}
\forall I^{i} \in C', \ U_{max}(I^{i}, C') > U_{max}(I^{i}, C^{j}) \ (I^{i} \in C^{j} \in S).
\label{eq:ce2}
\end{equation}

Suppose that $C^{0}$ denotes the coalition in the first iteration. Then $\forall I^{i} \in C^{0}$, $I^{i}$ achieves its maximal utility in the client set $\bm{I}$ and cannot increase its model utility further. So we have

\begin{equation}
\forall I^{i} \in C', \ I^{i} \notin C^{0}.
\end{equation}

The clients in the coalition $C^{0} \in S$ identified in the first iteration have no interest in collaborating with others and will be removed. Suppose $C^{1}$ denotes the coalition in the first iteration. $\forall I^{i} \in C^{1}$, $I^{i}$ achieves its maximal utility in the client set $\bm{I} \backslash C^{0}$ and cannot increase its model utility further. So we have

\begin{equation}
\forall I^{i} \in C', \ I^{i} \notin C^{1}.
\end{equation}

Then $C^{1}$ will be removed and we have $\forall I^{i} \in C', \ I^{i} \notin I^{2}.$. Based on the same analysis, we finally have
\begin{equation}
\begin{aligned}
& \forall I^{i} \in C', \ \forall C^{j} \in S, \ I^{i} \notin C^{j}, \\
& \Rightarrow C' = \emptyset.
\end{aligned}
\end{equation}

Therefore, we prove that there is no nonempty $C'$ satisfying Eq.(\ref{eq:ce2}) and $S$ satisfies Outer Agreement.
\end{proof}

\subsection{Proof of Proposition 2}
\begin{proof}
For any hypothesis $h'^{*} \in P(C')$, $h'^{*}$ satisfies

\begin{equation}
\nexists h^{\prime} \in \mathcal{H}, \text { s.t. } \forall i \in C': l_{i}\left(h^{\prime}\right) \leq l_{i}(h'^{*}) \text { and } \exists j: l_{j}\left(h^{\prime}\right)<l_{j}(h'^{*}).
\label{eq:prop1}
\end{equation}


1) If $h^{*} \in P(C)$ which satisfies

\begin{equation}
\nexists h^{\prime} \in \mathcal{H}, \text { s.t. } \forall i \in C: l_{i}\left(h^{\prime}\right) \leq l_{i}(h) \text { and } \exists j: l_{j}\left(h^{\prime}\right)<l_{j}(h),
\label{eq:prop3}
\end{equation}

then we let $h^{*} = h'^{*}$ in Eq.(\ref{eq:prop}) so Eq.(\ref{eq:prop}) holds;

2) if $h'^{*} \notin P(C)$, there exists $h' \in P(C)$ which satisfies

\begin{equation}
\forall i \in C: l_{i}\left(h^{\prime}\right) \leq l_{i}(h'^{*}) \text { and } \exists j: l_{j}\left(h^{\prime}\right)<l_{j}(h'^{*}),
\label{eq:prop4}
\end{equation}

combining Eq.(\ref{eq:prop1}), we have

\begin{equation}
\forall i \in C': l_{i}\left(h^{\prime}\right) = l_{i}(h'^{*}) \text { and } \exists j \in C \backslash C': l_{j}\left(h^{\prime}\right)<l_{j}(h'^{*}).
\end{equation}
then we let $h^{*} = h'$ in Eq.(\ref{eq:prop}) so Eq.(\ref{eq:prop}) holds.
\end{proof}

\section{Implementation Details}

\subsection{Learning Pareto Front}
Multi-objective optimization (MOO) problems have a set of optimal solutions, and these optimal solutions form the Pareto front, where each point on the front represents a different trade-off among possibly conflicting objectives. We construct a personalized model for each client which we call target network and it has the same architecture as the baselines. To determine the parameters of the target network of each client, in our experiments, we learn the entire Pareto Front simultaneously using a hypernetwork which receives as input a vector $\bm{d}$ and returns all parameters of the target network. The input vector is N-dimension $\bm{d}$ in which each entry $d^{i}$ corresponds to the client $I^{i}$ and is sampled from the convex hull $\mathcal{D}$,
\begin{equation}
\mathcal{D} = \left\{ d, | \forall i, 1\leq i \leq N, d^{i} \geq 0, \ \text{and} \ \sum_{i = 1}^{N} d^{i} = 1 \right\}.
\end{equation}
Specifically, we construct the hypernetwork following in the architecture introduced in ~\citep{navon2020learning}. By a n-layer ($n \leq 3$) MLP network with the activation function ReLU, we infer the parameters of the target network. Then the obtained parameters will be assigned to the target network. We evaluate the performance of the parameters on the training set and the gradient information will be returned for updating the hypernetwork.

\subsection{Identifying the Optimal Collaborator Set}
From the statement above, we train a hypernetwork $HN$ for learning the whole Pareto front. $HN$ bridges the mapping from the vector $\bm{d}$ to the corresponding model parameters of the target network. To obtain the optimal target network parameters that achieve the minimal value loss of the target objective on the validation set, we optimize the vector $\bm{d}$ by gradient descent. Specifically, given an initial direction $d_{0}$, we firstly obtain the parameters of the target network using the learned hypernetwork $HN(d_{0})$. Like training the hypernetwork, the obtained parameters will be assigned to the target network. Then we evaluate the performance of the generated parameters on the validation set and compute the gradient of the input direction $d_{0}$. Finally, the gradient information will be used for updating the vector $\bm{d}$ iteratively until convergence.
\begin{equation}
\begin{aligned}
& d_{i+1} = d_{i} - \eta \cdot \nabla_{d_{i}} l^{*} \\
& d_{i+1} \leftarrow Clip(d_{i+1}) \\
& d_{i+1} \leftarrow Normalization(d_{i+1}),
\end{aligned}
\end{equation}
where $\eta$ denotes the learning rate, $Clip(d_{i+1})$ means that we clip each values $ d^{j}_{i+1} \in d_{i+1}$ to satisfy $\epsilon_{0} \leq d^{j}_{i+1} \leq 1 $ and $Normalization(d_{i+1})$ is as follows,
\begin{equation}
\begin{aligned}
\forall j\ (1 \leq j \leq N),  d'^{j}_{i+1} = \frac{d^{j}_{i+1}}{\sum_{j = 1}^{N} d^{j}_{i+1}}.
\end{aligned}
\end{equation}
Finally, we reach the optimal direction $d^{*}$ and its corresponding Pareto model $M^{*} = HN(d^{*})$.

For each client, we determine the optimal collaborator set by the optimal direction $d^{*}$ and the loss value of all objectives $\bm{l} = [l_{1}(M^{*}), l_{2}(M^{*}), ..., l_{N}(M^{*})]$. From the Pareto Front Embedded property described in Proposition~\ref{prop:embed-pt}, an ideal direction $d^{*}$ is a sparse vector in which the indexes of the non-zero values in $d^{*}$ are the necessary collaborators for the target client. In our experiments, $d^{*}$ usually is not a sparse vector, but there are values in $d^{*}$ which are significantly small. Therefore, we set a threshold $\epsilon$ for $d^{*}$ to determine which clients are necessary/unnecessary.

\subsection{Datasets and Training Resources}
In our experiments, UCI adult~\cite{kohavi1996scaling} and CIFAR10~\cite{krizhevsky2009learning} are public datasets. eICU~~\cite{pollard2018eicu} is a dataset for which permission is required. We followed the procedure on the website \url{https://eicu-crd.mit.edu} and got the approval to the dataset.

We run our experiments on a local Linux server that has two physical CPU chips (Inter(R) Xeon(R) CPU E5-2640 v4 @ 2.40GHz) and 32 logical kernels. We use Pytorch framework to implement our model and train all models on GeForce RTX 2080 Ti GPUs.

\end{document}